\documentclass[letterpaper, 10 pt, conference]{ieeeconf}

\IEEEoverridecommandlockouts

\usepackage{dblfloatfix}

\usepackage{color, colortbl}
\usepackage[colorlinks,bookmarksnumbered,citecolor=orange,urlcolor=orange]{hyperref}
\usepackage{graphicx}
\usepackage{svg}
\graphicspath{{Figures}}
\DeclareGraphicsExtensions{.pdf,.png}
\usepackage{bigstrut}
\usepackage[english]{babel}

\usepackage{csquotes}

\usepackage[T1]{fontenc}
\usepackage{url}
\usepackage{multirow}
\usepackage{float}
\usepackage{xcolor}
\usepackage{hyperref}
\hypersetup{
     colorlinks=true,
     linkcolor=orange,
     filecolor=orange,
     citecolor=orange,      
     urlcolor=orange,
     }

\usepackage[table]{xcolor}

\let\labelindent\relax
\usepackage{graphicx}
\usepackage{amsmath}
\usepackage{enumitem}
\usepackage{svg}
\usepackage{float}   
\usepackage[ruled,vlined]{algorithm2e}
\usepackage{epstopdf}

\usepackage{graphics} 
\usepackage{epsfig} 
\usepackage{times} 
\usepackage{amssymb}  
\usepackage{xcolor}

\usepackage{booktabs}
\usepackage{pifont}

\usepackage{cite} 

\usepackage{glossaries}
\usepackage{multicol}
\usepackage{comment}

\title{\LARGE \bf
Return-to-Home Feasible Micro-Aerial Vehicle Exploration for 3D Gaussian Splatting Reconstruction
}

\author{
     Prajit KrisshnaKumar, Fan Yang, and Koichiro Niinuma
    \thanks{All authors are affiliated with Fujitsu Research.}
    \thanks {\copyright\space2026 IEEE. Personal use of this material is permitted. Permission from IEEE must be obtained for all other uses, in any current or future media, including reprinting/republishing this material for advertising or promotional purposes, creating new collective works, for resale or redistribution to servers or lists, or reuse of any copyrighted component of this work in other works.}
}

\begin{document}

\maketitle
\thispagestyle{empty}
\pagestyle{empty}

\begin{abstract}
Micro aerial vehicles (MAVs) enable rapid indoor 3D reconstruction for inspection and time-critical situational awareness, but must operate under strict flight-time budgets and safety constraints that require explicit return-to-home (RTH) feasibility with a conservative margin. We present a topology-first active reconstruction framework that decouples high-fidelity rendering from navigation. A dense 3D Gaussian Splatting (3DGS) map is optimized as the reconstruction target, while a lightweight Truncated Signed Distance Field (TSDF)/occupancy scaffold supports conservative collision checking and online construction of a sparse 3D Voronoi skeleton roadmap. Since directly extracting roadmaps from TSDF geometry can be unstable under noisy and incomplete online fusion, we validate nodes and edges using carved free-space consistency and dense visibility checks, which suppress behind-wall phantom structure and stabilize planning. Viewpoints are selected on the roadmap using a flight-time-budget-aware objective and a lightweight receding-horizon lookahead to improve non-myopic exploration behavior. We evaluate on photorealistic indoor simulation benchmarks (ReplicaCAD, Gibson, and HM3D), reporting reconstruction coverage/error, RTH success, and computational cost. Results demonstrate improved or competitive performance relative to recent GS-based baselines under identical flight-time budgets.
\end{abstract}

\section{Introduction}

\begin{figure}
\centering
\includegraphics[width=\linewidth]{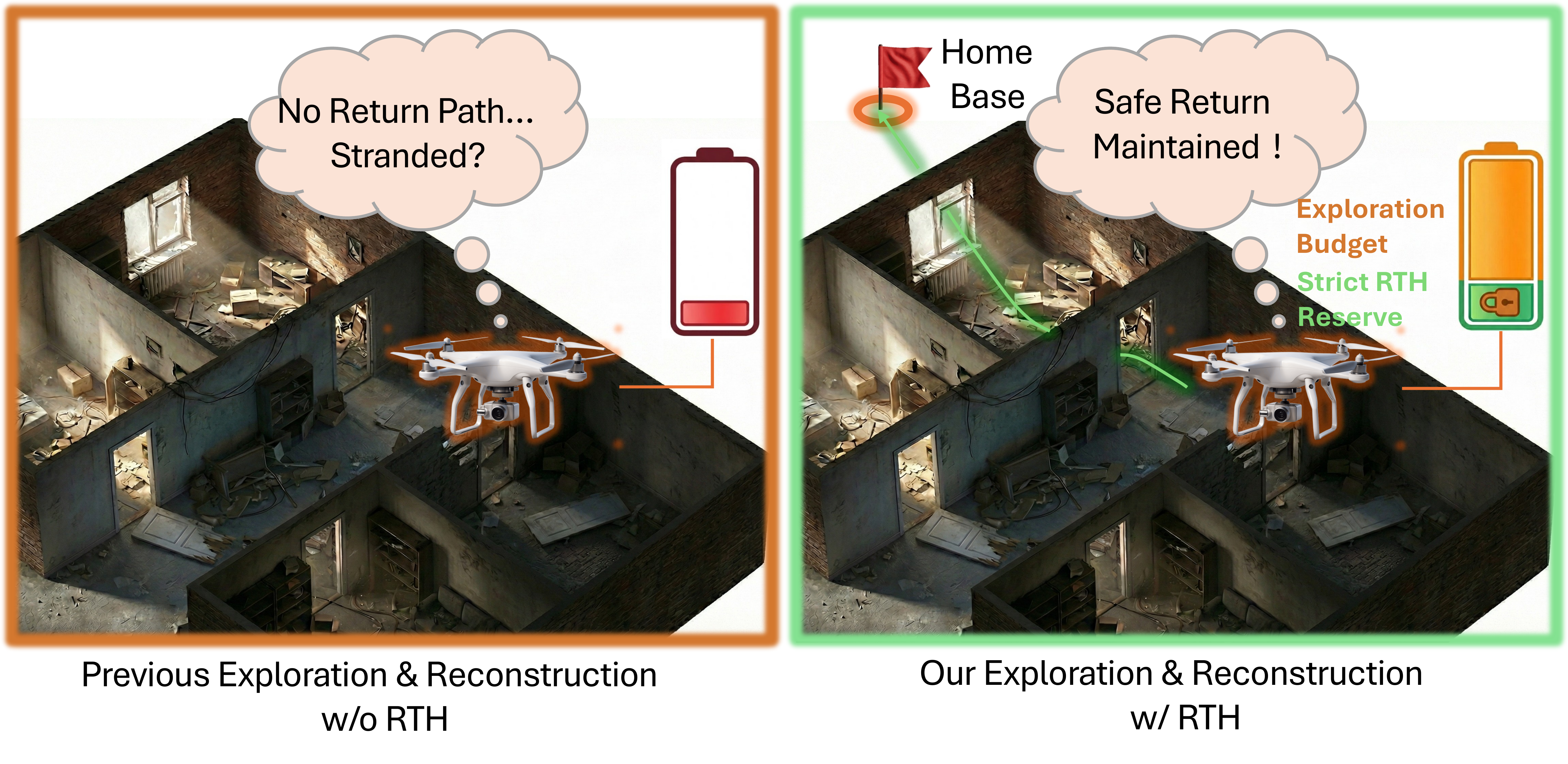}
\vspace{-2em}
\caption{Exploration and reconstruction of hazardous buildings must explicitly account for the return-to-home (RTH) constraint to ensure safe MAV recovery.}
\label{fig:demo}
\vspace{-0.4cm}
\end{figure}

Micro aerial vehicles (MAVs) are increasingly attractive for indoor search-and-rescue (SAR) and first-responder missions because they can quickly enter confined or hazardous spaces and provide situational awareness before humans are deployed \cite{kumar2012opportunities}.
In these environments, autonomy must handle cluttered, GPS-denied navigation through narrow doorways and corridors under strict flight-time limits. 
If the MAV exhausts its battery deep inside a structure, safe recovery is not guaranteed and the mission may fail prematurely.
Consequently, a central operational requirement is \emph{return-to-home} (RTH) feasibility. In multi-room indoor layouts, return cost is highly non-Euclidean, and straight-line distance can severely underestimate the true time required to navigate around walls and through doorways. 
Therefore, viewpoint selection must trade off expected reconstruction gain against travel time while maintaining a conservative safety margin for possible interruptions during the return flight (Fig.~\ref{fig:demo}).

Beyond geometric coverage, responders often require \emph{high-fidelity} reconstructions that preserve appearance and
fine structure for inspection and decision making. 
3D Gaussian Splatting (3DGS)~\cite{kerbl20233dgs} enables photorealistic reconstruction by representing scenes with explicit Gaussian primitives and fast differentiable rasterization, making online refinement feasible at interactive rates. 
This has motivated active reconstruction systems that couple viewpoint planning with online map updates. 
However, bringing 3DGS-based active reconstruction to MAVs is challenging because informative viewpoint selection must remain \emph{execution-aware} under the above non-Euclidean travel and RTH constraints, and because a 3DGS map is rendering-centric and does not directly provide watertight surfaces or conservative free-space semantics for collision checking. 
Practical autonomy therefore typically requires an auxiliary geometric scaffold for safe navigation and feasibility reasoning.

While recent work has begun coupling Gaussian splatting with active viewpoint planning, reliably integrating high-fidelity mapping with an execution-ready navigation abstraction under budget constraints in multi-room indoor scenes remains challenging. 
ActiveGS~\cite{jin2025activegs} uses a voxel scaffold to represent unexplored space and guide exploration, which can become costly to maintain and query as the explored 3D volume grows. ActiveSplat~\cite{li2025activesplat} enables efficient planning using a planar topological abstraction, but this 2D workspace does not explicitly capture the 3D connectivity and occlusions that arise in indoor flight.

To bridge these gaps, we take a topology-first approach that \emph{decouples} photorealistic reconstruction from conservative navigation reasoning. 
The MAV incrementally optimizes a dense 3DGS map as the reconstruction output, while simultaneously fusing the same posed RGB-D observations into a lightweight TSDF and occupancy grid that provide explicit surfaces and conservative free-space evidence for planning.
From this evolving geometric scaffold, we construct a sparse \emph{3D Voronoi skeleton roadmap} whose vertices lie near the medial axis of free space and naturally provide high-clearance waypoint candidates, similar in spirit to classical Voronoi-graph roadmaps used for sensor-based exploration~\cite{choset2000sensor}. 
Since online TSDF geometry is incomplete early in exploration and can contain holes and transient artifacts, directly extracting a roadmap can introduce behind-wall phantom nodes that waste flight time and jeopardize safe return. We therefore apply conservative validation using carved free-space consistency and dense visibility checks to keep the roadmap stable for MAV execution.
On the resulting roadmap, we formulate viewpoint actions as (node, yaw) pairs and select goals using a budget-aware objective that trades off information gain against \emph{shortest-path} travel time while enforcing an explicit RTH safety margin. 
Finally, to reduce myopic oscillations, we use a lightweight two-step receding-horizon
selection, motivated by receding-horizon next-best-view planning for aerial exploration~\cite{bircher2016receding}.

In summary, our main contributions are as follows:

\begin{itemize}
    \item We introduce a drone-centric hybrid planning architecture that decouples photorealistic reconstruction from conservative navigation: a dense 3DGS map is optimized as the reconstruction target, while a lightweight TSDF occupancy scaffold enables collision checking and \emph{stable} online extraction of a sparse 3D Voronoi skeleton roadmap via carved-free-space validation, behind-wall suppression, and free-space node injection.
    \item     We propose a budget-aware flight-time mechanism that couples structural awareness (information gain and frontier cues) with a travel-cost model. The resulting objective explicitly enforces a return-to-home safety margin and adaptively trades off information gain against travel time as remaining flight-time budget decreases.
    \item We provide evaluation under continuous-time MAV execution in photorealistic simulation, including comparisons to recent GS-based active reconstruction baselines and ablations isolating the effect of roadmap construction and non-myopic selection.
\end{itemize}

\section{Related Work}
\paragraph{Autonomous Exploration and Informative Path Planning}
Informative Path Planning (IPP) selects sensing actions to maximize information gain or coverage under constraints such as time, range, or energy~\cite{zhu2021online}. In unknown environments, IPP is commonly solved online using an evolving map so that viewpoint selection adapts to newly observed structure. For UAV exploration in large 3D spaces, scalable representations and clearance-aware planning are important; examples include low-memory mapping for exploration~\cite{huang2024efficient} and Voronoi-guided planning with adaptive layering~\cite{dong2025voronoi}. Complementary work targets robust state estimation and globally consistent geometric mapping, such as GO-SLAM~\cite{zhang2023go}, Voxgraph~\cite{reijgwart2019voxgraph}, and Kimera~\cite{rosinol2020kimera}.

\paragraph{High-Fidelity Scene Reconstruction}
NeRF-based methods produce high-quality view synthesis but rely on expensive volume rendering, which limits real-time closed-loop use. 3D Gaussian Splatting (3DGS)~\cite{kerbl20233dgs} provides an efficient alternative via explicit
Gaussian primitives and fast rasterization, enabling interactive rendering and online refinement and motivating NBV-style active reconstruction~\cite{bircher2016receding,charrow2015information}. Recent systems estimate view utility from renderability or uncertainty cues~\cite{chen2025activegamer,xie2025gaussmi,li2025active3d}. Since 3DGS is rendering-centric and does not provide conservative free-space semantics, many approaches pair it with auxiliary geometric scaffolds or sparse navigation abstractions for safe planning, such as ActiveGS~\cite{jin2025activegs}, ActiveSplat~\cite{li2025activesplat}, and neural-map exploration methods~\cite{kuang2024active}.

\paragraph{UAV-Based Indoor Reconstruction and Unresolved Gaps}
Indoor UAV reconstruction systems study exploration and mapping under practical sensing and execution constraints~\cite{zhang2025autonomous,wu2024optimally}. Recent 3DGS-focused planners also target quadrotor autonomy, including GS-Planner~\cite{jin2024gs} and HGS-Planner~\cite{xu2024hgs}. Despite progress, robust autonomous reconstruction in cluttered multi-room indoor scenes remains challenging. Dense voxel scaffolds for unknown-space reasoning can become expensive as explored volume grows, while planar abstractions can miss 3D connectivity and occlusions important for indoor flight. Moreover, return-to-home feasibility is inherently non-Euclidean indoors because return cost must respect doors and walls, and sparse roadmaps extracted from online fusion can become unstable early in exploration without conservative validation. These limitations motivate hybrid architectures that optimize 3DGS as the reconstruction target while planning and safety reasoning are performed on a lightweight geometric scaffold under explicit budget and RTH feasibility constraints.

\section{Methodology}
\label{sec:methodology}
\begin{figure*}
    \centering
    \includegraphics[width=1\linewidth]{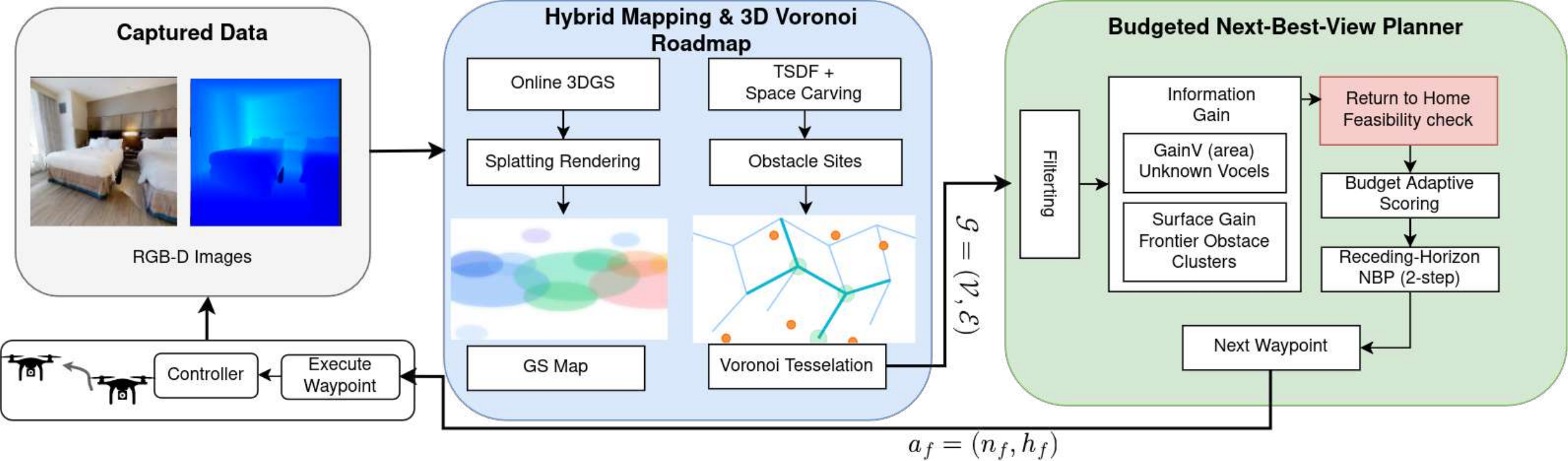}
    \vspace{-1em}
    \caption{\textbf{System Overview}. Our proposed framework integrates Online 3D Gaussian Splatting (3DGS) for high-fidelity reconstruction with a Voronoi-based roadmap for efficient navigation. The Budgeted Next-Best-View (NBV) Planner dynamically balances information gain against remaining budget to ensure feasible, long-horizon exploration.}
    \label{fig:framework}
    \vspace{-0.5cm}
\end{figure*}

\subsection{Method Overview}

We address edge-assisted budgeted autonomous exploration for high-fidelity indoor reconstruction using an RGB-D equipped micro aerial vehicle. A fundamental challenge in active radiance-field mapping is that representations optimized for photorealistic rendering (e.g., 3DGS) lack the explicit, conservative free-space semantics required for safe navigation. To resolve this, our framework (Fig.~\ref{fig:framework}) adopts a hybrid perception--action loop. Specifically, the system: (i) incrementally optimizes a dense 3DGS map as the primary reconstruction target, while concurrently (ii) fusing depth into a lightweight TSDF to serve as a geometric scaffold for extracting a sparse 3D topological roadmap. Using this roadmap, (iii) a budget-aware planner selects and executes next-best viewpoints to maximize coverage while enforcing return-to-home (RTH) feasibility.

\subsection{Gaussian Splatting}
We represent the scene as a set of explicit 3D Gaussian primitives \cite{kerbl20233dgs}.
Each Gaussian $i$ is parameterized by a mean $\boldsymbol{\mu}_i$, an
anisotropic covariance $\boldsymbol{\Sigma}_i$, a color $\mathbf{c}_i$,
and an opacity $o_i$.
Given a camera pose, the per-pixel contribution of Gaussian $i$ at pixel
$\mathbf{u}$ can be written as
\begin{equation}
\alpha_i(\mathbf{u}) =
o_i \exp\!\left(
-\tfrac{1}{2}\big(\mathbf{x}(\mathbf{u})-\boldsymbol{\mu}_i\big)^\top
\boldsymbol{\Sigma}_i^{-1}
\big(\mathbf{x}(\mathbf{u})-\boldsymbol{\mu}_i\big)
\right),
\end{equation}
where $\mathbf{x}(\mathbf{u})$ denotes the 3D point associated with pixel
$\mathbf{u}$ in the camera coordinate frame.
Rendered color is obtained by standard front-to-back $\alpha$-compositing:
\begin{equation}
\hat{\mathbf{C}}(\mathbf{u}) =
\sum_{i} \mathbf{c}_i\,\alpha_i(\mathbf{u})
\prod_{j<i}\bigl(1-\alpha_j(\mathbf{u})\bigr),
\end{equation}
with Gaussians ordered by increasing depth.
Depth and accumulated opacity are computed analogously using the same compositing rule (inherited from \cite{li2025activesplat}). 

\subsection{3D Voronoi Roadmap Construction}\label{subsec:voronoi_roadmap}
We construct a sparse, topologically meaningful abstraction of the environment: a 3D Voronoi roadmap whose nodes approximate the medial axis of free space and whose edges maximize obstacle clearance. To ensure the roadmap remains stable under continuous map updates, we employ a pipeline that fuses depth into a TSDF, extracts temporally smoothed obstacle sites, computes a 3D Voronoi tessellation, and rigorously filters the resulting graph using explicit volumetric carving.

\subsubsection{Volumetric Fusion and Space Carving}
At each timestep $k$, the drone captures a posed RGB-D observation, denoted as the tuple $(I_k, D_k, \mathbf{T}_{W\leftarrow C,k})$. Here, $I_k$ is the RGB image, $D_k$ is the aligned depth map, and $\mathbf{T}_{W\leftarrow C,k} \in SE(3)$ is the rigid camera pose mapping points from the camera frame $C$ to the world frame $W$. Using the camera's known intrinsic parameters, we back-project each pixel $\mathbf{u}$ from the depth map $D_k$ to obtain a 3D point cloud $\mathcal{P}_k$ in world coordinates:
\begin{equation}
\mathbf{p}_{W}(\mathbf{u}) = \mathbf{T}_{W\leftarrow C,k}\, \pi^{-1}\!\big(\mathbf{u}, D_k(\mathbf{u})\big),
\end{equation}
where $\pi^{-1}(\cdot)$ denotes the pinhole back-projection function. 

Often, these depth measurements contain noise, incomplete information, floating outliers, and view-dependent missing regions. To obtain a stable geometric estimate suitable for topology extraction, we apply voxel-grid downsampling to remove duplicates and reduce fusion cost. We then integrate this filtered point cloud into a sparse volumetric TSDF \cite{curless1996volumetric}. The TSDF stores a signed distance $\phi(\mathbf{x})$ and weight $w(\mathbf{x})$ per voxel center $\mathbf{x}$, which are updated iteratively as a weighted average of the truncated distance observations.

In addition to the TSDF surface, we maintain a conservative notion of free versus occupied space for fast wall-crossing checks \cite{hornung2013octomap}. For each depth endpoint $\mathbf{p}_{W,k}(\mathbf{u})$, we ray-march from the camera origin $\mathbf{o}_k$ to the endpoint. Voxels along the ray before the endpoint are marked \emph{free}, and the endpoint voxel is marked \emph{occupied}. Let $\mathcal{F}$ be the set of free voxels and $\mathcal{O}$ the set of occupied voxels at a fixed resolution $r$ (e.g., $r=0.20$\,m). Because noisy depth endpoints can temporarily mark a previously carved free voxel as occupied (creating phantom geometry or "floaters"), we define an effective occupied set using $\mathcal{O}_{\mathrm{eff}} = \mathcal{O}\setminus \mathcal{F}$
which prevents these ``disputed'' voxels from incorrectly blocking doorways or corridors.

\subsubsection{Obstacle Sites}
We extract a surface mesh from the TSDF zero-level set using marching cubes~\cite{lorensen1998marching}. The Voronoi tessellation is computed from a discrete set of generator points; if these generators change slightly between cycles, Voronoi vertices and edges can change disproportionately.
To reduce this sensitivity, we convert the extracted mesh into a persistent set of obstacle \emph{sites}
$\mathcal{S}=\{(\mathbf{s}_i,\mathbf{n}_i)\}$ with associated surface normals.
Sites are stored in a coarse voxel-hash map and updated with an exponential moving average (EMA):
\begin{equation}
\mathbf{s}_i \leftarrow (1-\eta)\mathbf{s}_i + \eta\,\tilde{\mathbf{s}}_i,\qquad
\mathbf{n}_i \leftarrow \frac{(1-\eta)\mathbf{n}_i + \eta\,\tilde{\mathbf{n}}_i}{\|(1-\eta)\mathbf{n}_i + \eta\,\tilde{\mathbf{n}}_i\|},
\label{eq:ema_sites}
\end{equation}
where $(\tilde{\mathbf{s}}_i,\tilde{\mathbf{n}}_i)$ denotes the current cycle's measurement in the corresponding hash cell and $\eta\in(0,1)$ is a small smoothing factor. This update acts as a spatial low-pass filter and reduces roadmap jitter under incremental map updates.

\subsubsection{Voronoi Tessellation}\label{subsec:voronoitessellation}
Given obstacle site positions $\{\mathbf{s}_i\}$, the Voronoi cell of site $\mathbf{s}_i$ is
\begin{equation}
\mathcal{V}_i=\left\{\mathbf{x}\in\mathbb{R}^3 \;\middle|\;
\|\mathbf{x}-\mathbf{s}_i\|\le \|\mathbf{x}-\mathbf{s}_j\|,\ \forall j \right\}.
\label{eq:voronoi_cell}
\end{equation}
Finite Voronoi vertices (intersections of multiple cells) are equidistant to several nearby obstacle sites and therefore lie near the medial axis of free space.
We take all finite Voronoi vertices as candidate nodes and attach a clearance attribute
\begin{equation}
\mathrm{clr}(\mathbf{v})=\min_i \|\mathbf{v}-\mathbf{s}_i\|,
\label{eq:clearance}
\end{equation}
computed by nearest-neighbor queries in the site set.
Candidate edges are induced by finite Voronoi ridges connecting pairs of surviving vertices where each edge is weighted by Euclidean length. These raw Voronoi vertices include many invalid candidates, such as points outside the explored volume, points inside walls, points in unobserved void, and points behind walls that appear only due to sparse sampling or holes.
To overcome this, we apply the below filters that each targets a specific failure mode. 

\begin{itemize}
\item \textbf{Minimum safety distance.}
We reject vertices with $\mathrm{clr}(\mathbf{v})<d_{\mathrm{safe}}$ to enforce a collision margin compatible with drone size and control uncertainty.
\item \textbf{Proximity to carved free space.}
We reject vertices that are not near any confirmed free voxel in $\mathcal{F}$ (within radius $r_F$), preventing the selection of nodes in unknown space that are not connected to exploration.
\item \textbf{Dense occupancy line-of-sight (behind-wall suppression).}
For vertices that still pass the above checks, we ray-march the segment from $\mathbf{v}$ to a few nearest camera centers through $\mathcal{O}_{\mathrm{eff}}$.
If all such rays intersect occupied voxels, the vertex is behind a wall relative to the explored free space and is removed \cite{amanatides1987fast}.
\end{itemize}
These filtering strategies prioritize removing unreachable ``phantom'' nodes over retaining every possible candidate near tight geometry; the TSDF resolution affects surface-site extraction, while traversability is ultimately checked through carved free-space, occupancy traversal, and the safety margin.

\subsubsection{Graph Sparsification}
Even if two vertices are individually valid, the straight segment between them may cross a wall.
For each candidate edge $e=(\mathbf{v}_a,\mathbf{v}_b)$ we sample points along the segment following $\mathbf{x}(t) = (1-t)\mathbf{v}_a + t\mathbf{v}_b, t\in(0,1)$
and reject the edge if either (i) any sample violates a minimum clearance threshold with respect to the obstacle sites, or (ii) any sample voxel falls in $\mathcal{O}_{\mathrm{eff}}$ (dense occupancy traversal).
Finally, we remove tiny disconnected components, sparsify the graph (e.g., contracting near-collinear degree-2 chains and merging nearby nodes to bound computation).

\subsubsection{Free-Space Node Injection and Bridging} \label{subsub:freespace}
\begin{figure}
    \centering
    \includegraphics[width=1\linewidth]{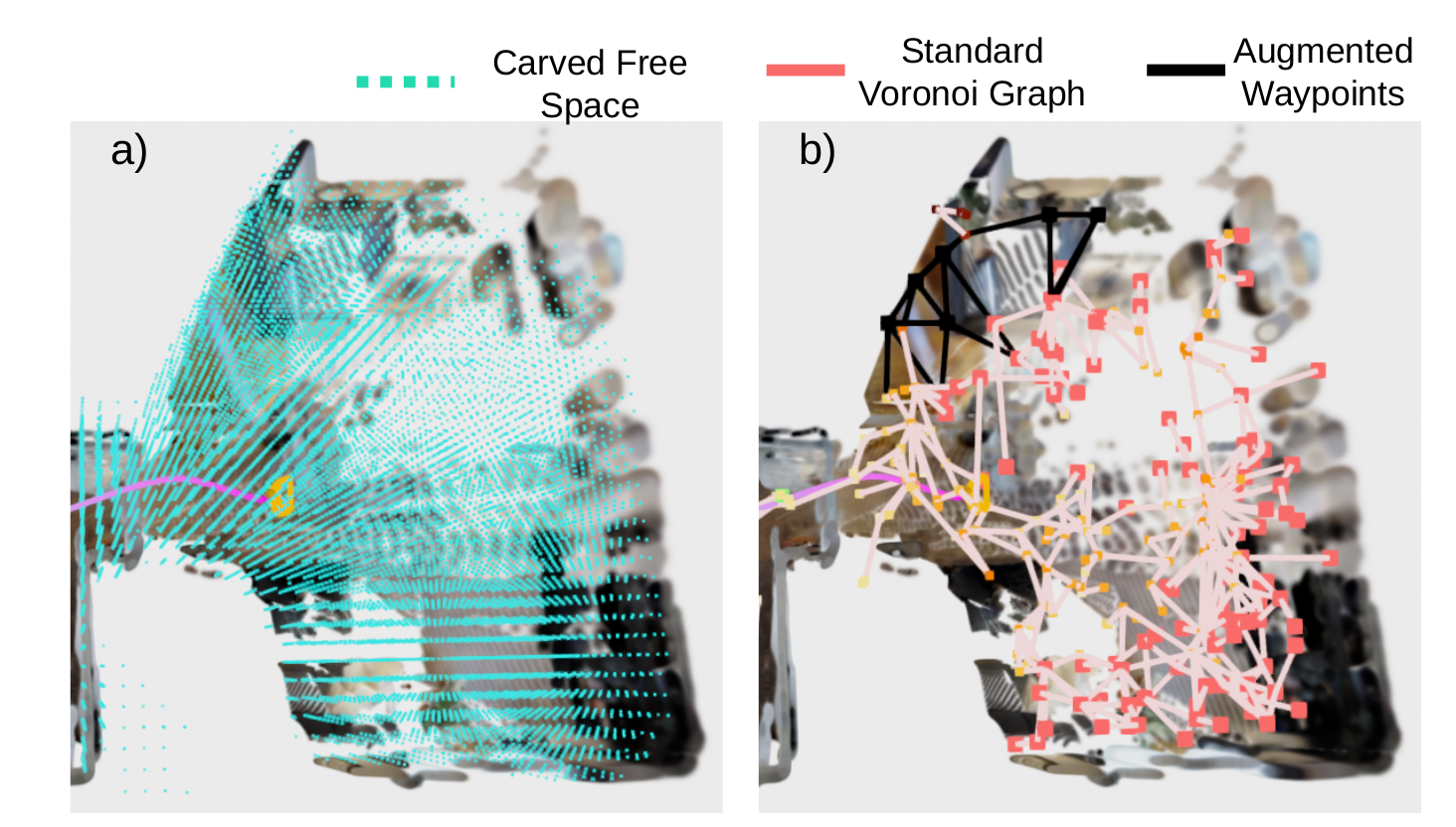}
    \vspace{-0.4cm}
    \caption{Free-space node injection. 
(a) Volumetric visualization of the explicitly carved free-space grid $\mathcal{F}$ in a wide-open environment. 
(b) The resulting 3D Voronoi roadmap, where orange segments show the raw Voronoi graph clustered near perimeter walls, and black segments show synthetically injected nodes and bridging edges that use $\mathcal{F}$ to safely reduce coverage gaps.}
    \label{fig:carved_freespace}
    \vspace{-0.4cm}
\end{figure}

The raw 3D Voronoi graph can under-sample wide-open regions because Voronoi vertices tend to concentrate near obstacle boundaries. To improve coverage and connectivity, we inject additional waypoint nodes directly from the carved free-space grid $\mathcal{F}$ and connect them to the existing roadmap using the same clearance and $\mathcal{O}_{\mathrm{eff}}$ collision checks as for standard edges. This augmentation fills coverage gaps in open areas and provides safe ``bridges'' between weakly connected components.
Figure~\ref{fig:carved_freespace} visualizes the carved free space and the resulting augmented roadmap.
The output of this stage is a sparse, collision-checked roadmap $\mathcal{G}=(\mathcal{V},\mathcal{E})$ with per-node attributes (position, clearance), which is then used for viewpoint scoring and planning in subsequent sections.

\subsection{Viewpoint Selection}
At each planning iteration, the planner requests a snapshot of the current 3D Voronoi roadmap $\mathcal{G}=(\mathcal{V},\mathcal{E})$ from the previous step. 
Each node $n\in\mathcal{V}$ stores its 3D position $\mathbf{p}_n$ and per-node attributes (e.g., clearance), and the snapshot also provides two per-view gain tensors: GainV scores $g_{\mathrm{area}}(n,h)$ and frontier-surface scores $g_{\mathrm{surf}}(n,h)$ for a discrete set of yaw headings $h\in\{0,\dots,H-1\}$ (we use $H=8$ headings, spaced by $45^\circ$). An action or the next waypoint is defined as $a=(n,h)$: moving to node $n$ and observing with heading $h$. 

\subsubsection{Return-to-Home Feasibility Filters}
To ensure safe recovery, we enforce a hard return-to-home (RTH) constraint: a candidate goal is admissible only if the MAV can reach it and still return to the takeoff location with a conservative time reserve. We compute shortest-path travel times on the roadmap $\mathcal{G}$ using two Dijkstra passes~\cite{jasika2012dijkstra}:
$c_{\mathrm{go}}(n)$ is the travel time from the current state to node $n$, and $c_{\mathrm{home}}(n)$ is the travel time from $n$ back to the home node.
Let $B$ denote the total flight-time budget and let $t$ be the elapsed flight time. We define the remaining budget
$T_{\mathrm{rem}} = B - t$ and require
\begin{equation}
c_{\mathrm{go}}(n) + c_{\mathrm{home}}(n) + T_{\mathrm{res}}
\;\le\; T_{\mathrm{rem}},
\label{eq:rth_budget}
\end{equation}
where $T_{\mathrm{res}}$ is a fixed safety reserve (time margin). 

\subsubsection{Frontier Indicator}
Each candidate node is assigned a frontier strength score derived from two signals: (i) an \emph{obstacle-frontier} count, computed by counting surface points that have only been observed from far camera ranges in the neighborhood of the node, and (ii) a  \emph{free-space frontier} count, computed from classic occupancy-grid frontiers within a search radius \cite{yamauchi1997frontier}.

\subsubsection{Information Gain}
We have primarily two scoring methods to calculate the potential information gain for selecting an action $a$. Area discovery gain, termed as GainV ($g_{\mathrm{area}}(n,h)$) estimates how much new space a depth camera would reveal from action $a=(n,h)$ by raycasting a camera frustum from $\mathbf{p}_n$ at heading $h$ through a voxel grid that classifies cells as free, occupied, or unknown. It is essentially the number of unknown voxels observed before occlusion. While this term prioritizes new area, another term, frontier surface gain ($g_{\mathrm{surf}}(n,h)$ ) improves the reconstruction quality. This is achieved through clustering frontier obstacle points and scoring clusters that (i) are visible from the node (line-of-sight in the occupancy grid) and (ii) fall inside the view's field of view. 

\subsubsection{Objective Function}

We score each action $a=(n,h)$ using
\begin{equation}
U(n,h) = G(n,h)\cdot P(n),
\label{eq:score_c}
\end{equation}
where the gain term uses log-compression to stabilize scale:
\begin{equation}
\footnotesize
G(n,h)=\alpha(t)\,\ln(1+g_{\mathrm{area}}(n,h)) + (1-\alpha(t))\,\ln(1+g_{\mathrm{surf}}(n,h)).
\label{eq:gain_mix}
\end{equation}
We use a linear schedule $\alpha(t)=\mathrm{clip}(1 - t/B,\,0.2,\,0.9)$ where $B$ is the total flight-time budget.
It decays from a high value early in the mission (prioritizing rapid area discovery) to a lower value late in the mission (prioritizing refinement of incomplete surfaces). Together with the budget-adaptive travel penalty $P(n)$, this induces a natural behavior where the MAV expands aggressively early on and gradually contracts its operational radius as the remaining budget decreases, facilitating safe return while refining nearby geometry.

The travel penalty is an exponential discount in travel time:
\begin{equation}
P(n)=\exp(-\lambda_k\,c_{\mathrm{go}}(n)),\qquad \lambda_k=\frac{\ln 2}{\tau_k}.
\label{eq:travel_penalty}
\end{equation}
Crucially, rather than using a fixed penalty rate, we parameterize the decay using a budget-adaptive half-life $\tau_k$. Let $T_{\mathrm{rem}} = B - t$ denote the remaining flight-time budget. We set:
\begin{equation}
\tau_k = \max\!\left(\beta_k \cdot \frac{T_{\mathrm{rem}}}{2},\; \tau_{\mathrm{floor}}\right),
\label{eq:tau_k}
\end{equation}
where $k\in\{\text{frontier},\text{non-frontier}\}$ selects $\beta_k$ based on whether the node lies on an exploration frontier (we use $\beta_{\text{frontier}}=1$ and $\beta_{\text{non-frontier}}=\tfrac{1}{3}$), and $\tau_{\mathrm{floor}}$ is a small positive constant. This dynamic scaling makes long-distance travel competitive early in the flight but increasingly suppresses distant goals as the battery depletes.

\subsubsection{Receding-Horizon Next-Best-View (NBP) Search}\label{subsub:nbplite}
A purely greedy selection strategy (GainMax) that optimizes only $U(n,h)$ suffers from information myopia \cite{bircher2016receding}. It frequently drives the MAV into dead ends or isolated corners that offer a momentary spike in gain, forcing the drone to waste significant flight time backtracking to the main frontier.
To prevent the MAV from falling into these local information maxima, we employ a lightweight two-step receding-horizon search. Let $\mathcal{A}_K$ be the top-$K$ actions ranked by $U(n,h)$. For each first-step action $a_1=(n_1,h_1)\in\mathcal{A}_K$ and potential second-step action $a_2=(n_2,h_2)\in\mathcal{A}_K$, we evaluate their joint utility:
\begin{equation}
\footnotesize
J(a_1,a_2) = U(a_1) + \gamma\,U(a_2)\,\exp(-\lambda_{\mathrm{step}}\,c(n_1\rightarrow n_2))\,(1-\beta\,\mathrm{overlap}),
\label{eq:nbp_lite}
\end{equation}
where $c(n_1\rightarrow n_2)$ is the Dijkstra travel time between the nodes. We parameterize the second-step discount using a half-life, $\lambda_{\mathrm{step}}=\ln 2/\tau_{\mathrm{step}}$, with $\tau_{\mathrm{step}}=20$\,s (Table~I).
We use a cheap redundancy indicator $\mathrm{overlap}=\mathbb{I}\!\left[\|\mathbf{p}_{n_1}-\mathbf{p}_{n_2}\|<r_{\mathrm{ov}}\right]$ with $r_{\mathrm{ov}}=3.0$\,m, so that two-step pairs that remain in the same local region are down-weighted by $(1-\beta\,\mathrm{overlap})$.  We select the pair maximizing $J$ but execute only the first action $a_1$, replanning at the next iteration. This ensures the chosen immediate action not only provides high current gain, but also positions the drone favorably for a high-value subsequent observation, maintaining continuous forward momentum through the environment.

\vspace{-0.15cm}
\section{Experiments and Results}
\label{sec:experiments}
To validate the proposed exploration framework, we evaluate our MAV-based active reconstruction system in photorealistic indoor simulation to quantify reconstruction quality, safety (return-to-home feasibility), and computational cost. 

\subsection{Simulation Setup and Execution Model}
\label{sec:sim_setup}
All experiments are conducted in the Habitat simulator, which provides photorealistic indoor environments, synchronized RGB-D observations, and a simulation clock. We evaluate on datasets with increasing layout complexity: (i) compact apartment-style scenes (ReplicaCAD), and (ii) multi-room residential scenes with narrow doorways and long corridors (Gibson and HM3D).
For Gibson~\cite{xiazamirhe2018gibsonenv} we use the same scene set as prior work~\cite{yan2023active}, for ReplicaCAD~\cite{szot2021habitat} we select three one-bedroom apartments, and for HM3D~\cite{ramakrishnan2021hm3d} we evaluate on two scenes (HaxA7YrQdEC, TEEsavR23oF).
We target an edge-assisted MAV setting: the MAV executes sensing and waypoint control, while 3DGS optimization, roadmap construction, and viewpoint selection run on a workstation/base-station computer equipped with an NVIDIA RTX 3080 GPU and an AMD Ryzen 5950 CPU with 32 GB RAM. 
Planned actions are defined on the roadmap as a target node position and a discrete yaw heading (Sec.~\ref{subsub:nbplite}). The MAV executes the resulting roadmap path with a simple velocity controller (position P-control with a yaw-rate controller). To improve robustness in cluttered indoor layouts, we use a depth-triggered safety check that performs short recovery behaviors (e.g., skipping a blocked intermediate waypoint or aborting the current segment) before replanning. Prior to exploration, the MAV performs an initialization scan (a slow $360^\circ$ yaw rotation) at the start location to stabilize the initial map for all methods.
We use Habitat ground-truth poses and simulated RGB-D sensing to isolate viewpoint planning and mapping; thus, the reported RTH feasibility is map-relative and does not model VIO drift, real sensor noise, or aerodynamic disturbances.
Table \ref{tab:params} shows the key parameters used in our experiments. 
Roadmap travel times $c_{\mathrm{go}}(\cdot)$ and $c_{\mathrm{home}}(\cdot)$ are computed by Dijkstra on $\mathcal{G}$ with edge weights given by nominal translational time, $w(u,v)=\|\mathbf{p}_u-\mathbf{p}_v\|/v_{\mathrm{nom}}$ (we use $v_{\mathrm{nom}}=0.5$\,m/s).
We do not explicitly model acceleration/deceleration, short safety recoveries, or final in-place yaw alignment inside $c_{\mathrm{go}}/c_{\mathrm{home}}$; these effects are execution-dependent and are handled by the simulator clock and absorbed by the safety margin $T_{\mathrm{res}}$.

\begin{table}[t]
\centering
\footnotesize
\caption{Key parameters used in all experiments.}
\label{tab:params}
\begin{tabular}{l c}
\toprule
\textbf{Parameter} & \textbf{Value} \\
\midrule
Safety distance $d_{\mathrm{safe}}$ & 0.30 m \\
Free-space proximity radius $r_F$ & 0.12 m \\
TSDF voxel size $r_{\mathrm{tsdf}}$ & 0.30 m \\
Occupancy/Gain voxel size $r_{\mathrm{occ}}$ & 0.20 m \\
Roadmap update period & 2.0 s \\
NBP parameters $(K,\gamma,\tau_{\mathrm{step}},r_{\mathrm{ov}},\beta)$ &
$(15,\,0.7,\,20\text{ s},\,3.0\text{ m},\,0.3)$ \\
RTH safety margin $T_{\mathrm{res}}$ & $0.1B$ \\
\bottomrule
\vspace{-0.5cm}
\end{tabular}
\end{table}


\subsection{Baseline Comparison}
To evaluate the effectiveness of our proposed system, we compare it against state-of-the-art active 3DGS methods and a naive exploration strategy.  We do not include GS-Planner \cite{jin2024gs} or HGS-Planner \cite{xu2024hgs} because we were unable to obtain a runnable setup compatible with our evaluation environment and sensing/execution interface at submission time.
Across all experiments, the flight-time budget ($B$) varies by scene complexity: $80$\,s for ReplicaCAD, $120$\,s for Gibson scenes with $\leq 4$ rooms, $180$\,s for Gibson scenes with $\geq 6$ rooms, and $240$\,s for HM3D.

\subsubsection{ActiveSplat}
ActiveSplat~\cite{li2025activesplat} is a Gaussian-splatting-based active reconstruction system that plans over a sparse topological graph using \emph{discrete} motion primitives (e.g., \texttt{MOVE\_FORWARD} $d{=}6.5$\,cm and yaw turns of $\Delta\psi{=}10^\circ$, with optional pitch turns). To compare against our continuous-time MAV execution under a total flight time budget, we normalize ActiveSplat trajectories using a simple time proxy derived from the number of forward and turn actions (mapped to time using nominal translational and yaw rates in our simulator). Importantly, we do not modify ActiveSplat’s planner or objectives. We only align the sensing and actuation interface by using the same rigid, forward-facing RGB-D camera model for all methods.
To map ActiveSplat's discrete action sequence to a flight-time budget, we compute a proxy execution time $T_{\mathrm{proxy}}=N_F(d/v_{\mathrm{nom}})+N_T(\Delta\psi/\omega_{\mathrm{nom}})$,
where $N_F$ is the number of forward steps of length $d=0.065$\,m and $N_T$ is the number of yaw turns of size $\Delta\psi=10^\circ$.
We use the same nominal rates as our simulator controller, $v_{\mathrm{nom}}=0.5$\,m/s and $\omega_{\mathrm{nom}}$ such that a $10^\circ$ turn corresponds to $0.07$\,s.
We evaluate the baseline under the same flight-time budget $B$ by truncating the discrete sequence at the largest prefix satisfying $T_{\mathrm{proxy}}\le B$.

\subsubsection{ActiveGS}
ActiveGS \cite{jin2025activegs} selects viewpoints to improve Gaussian reconstruction under an execution model where motion cost is computed from Euclidean travel distance (i.e., $t \approx \|\Delta\mathbf{x}\|/v$) without explicitly modeling yaw rotation time, local collision avoidance, or recovery behaviors. We reproduce this baseline following its published motion-cost model and evaluate it under the same flight-time budget normalization described above. This baseline therefore serves as a reference for performance under an optimistic kinematic assumption, whereas our method executes viewpoints with a
physically constrained controller.



\subsubsection{Ours (with and without RTH constraint)}
\label{sec:baseline_ours}

We report two variants of our method: (i) \textbf{Ours (w/ RTH)}, which enforces the return-to-home feasibility constraint
(Eq.~\ref{eq:rth_budget}) with a fixed safety margin ($T_{res}$ set to $10\%$ of the total budget), and (ii) \textbf{Ours (w/o RTH)}, which disables the RTH filter.

\subsubsection{Metrics}
We evaluate geometric reconstruction via Coverage (Cov) and Error (Err) between point sets sampled from the ground-truth ($\mathcal{P}_{\mathrm{gt}}$) and reconstructed ($\mathcal{P}_{\mathrm{rec}}$) surfaces. Using the nearest-neighbor distance $d(p,\mathcal{P}_{\mathrm{rec}}) = \min_{q \in \mathcal{P}_{\mathrm{rec}}} \|p-q\|_2$ for $p \in \mathcal{P}_{\mathrm{gt}}$, Coverage is the percentage of ground-truth samples within a threshold $\delta$: $\mathrm{Cov} = \frac{100}{|\mathcal{P}_{\mathrm{gt}}|} \sum_{p} \mathbf{1}[d(p,\mathcal{P}_{\mathrm{rec}}) < \delta]$. Error is the mean distance: $\mathrm{Err} = \frac{100}{|\mathcal{P}_{\mathrm{gt}}|} \sum_{p} d(p,\mathcal{P}_{\mathrm{rec}})$, reported in cm. We set $\delta=5$\,cm for all experiments.

\begin{table}[htbp]
\centering
\footnotesize
\caption{Mean performance summary. Gibson/HM3D: \textbf{scene-level} means; ReplicaCAD: \textbf{object-level} mean. Cov (\%), Err(cm).}
\label{tab:baseline_results}
\vspace{-0.4em}
\resizebox{\columnwidth}{!}{%
\begin{tabular}{@{} l cc cc cc @{}}
\toprule
\multirow{2}{*}{\textbf{Method}} &
\multicolumn{2}{c}{\textbf{Gibson (scene)}} &
\multicolumn{2}{c}{\textbf{HM3D (scene)}} &
\multicolumn{2}{c}{\textbf{ReplicaCAD (object)}} \\
\cmidrule(lr){2-3}\cmidrule(lr){4-5}\cmidrule(lr){6-7}
 & \textbf{Cov.} & \textbf{Err.} & \textbf{Cov.} & \textbf{Err} & \textbf{Cov.} & \textbf{Err.} \\
\midrule
\textbf{ActiveGS}       & 70.4 & 13.12 & 47.5 & 27.50 & 42.9 & 10.90 \\
\textbf{ActiveSplat}    & 89.6 & 4.85  & 67.2 & 17.77 & 77.3 & 4.27 \\
\textbf{Ours (w/o RTH)} & 88.7 & 6.23  & 68.7 & 18.64 & 90.5 & 3.04 \\
\textbf{Ours (w/ RTH)}  & 88.8 & 6.27  & 66.2 & 18.14 & 91.0 & 2.93 \\
\bottomrule
\vspace{-0.3cm}
\end{tabular}
}
\end{table}

\subsubsection{Baseline Results}
Table~\ref{tab:baseline_results} summarizes mean reconstruction \emph{coverage} (Cov, $\uparrow$) and \emph{error} (Err, cm, $\downarrow$). Gibson and HM3D are reported as \emph{scene-level} means (averaged over scenes), while ReplicaCAD is reported as an \emph{object-level} mean (averaged over annotated object instances). We use object-level
evaluation for ReplicaCAD because it provides instance-level geometry and labels, and object coverage better reflects reconstruction fidelity in cluttered indoor environments where a MAV can exploit 3D viewpoints to observe occluded objects.  For methods executed through a continuous-time drone controller (ActiveGS and our variants), each setting is run five times and we report the mean.

\begin{figure*}[ht!]
    \centering
    \includegraphics[width=0.95\linewidth]{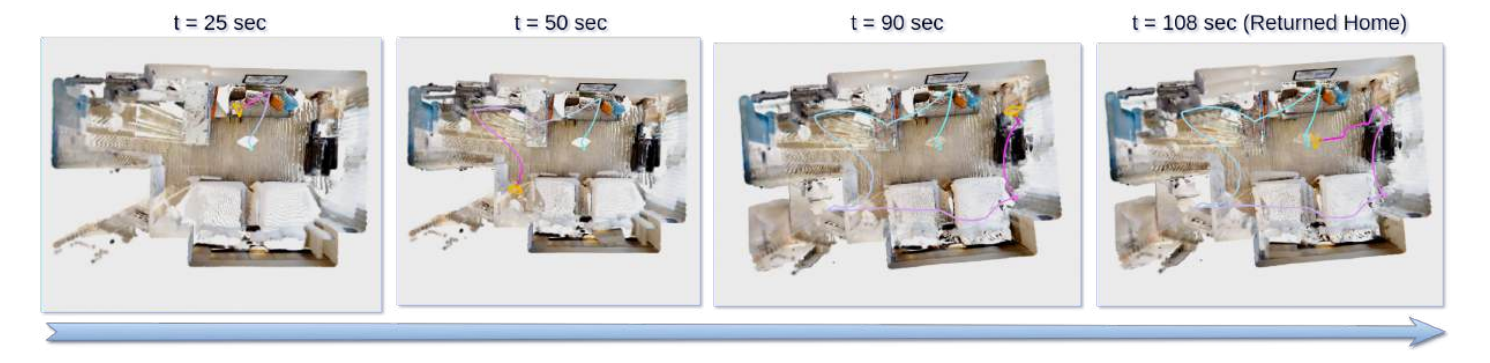}
    \vspace{-0.4cm}
    \caption{Qualitative 3DGS map snapshots over mission time, showing how the flight-time-aware planner first prioritizes rapid global exploration and later shifts toward local surface refinement as the time budget decreases.}
    \label{fig:qualitative_elmira}
    \vspace{-0.25cm}
\end{figure*}

ActiveSplat achieves the best mean performance on Gibson, which is consistent with its efficient planar abstraction working well in compact, largely single-floor apartment-style layouts. On HM3D, our no-RTH variant attains the highest mean coverage, and both of our variants remain competitive in error relative to ActiveSplat. On ReplicaCAD, our method improves both object coverage and error, highlighting the benefit of 3D viewpoints and clearance-aware planning for observing occluded objects beyond a purely ground-centric perspective. Finally, enforcing the RTH feasibility constraint has a small effect on reconstruction quality and can slightly improve it in some settings, since the return trajectory contributes additional observations and the feasibility constraint regularizes late-stage goal selection.
 Fig.~\ref{fig:qualitative_elmira} qualitatively illustrates this behavior: the MAV rapidly expands coverage early in the mission and then shifts toward refining remaining incomplete surfaces as the flight-time budget depletes.
ActiveGS is designed to optimize surface reconstruction quality using Gaussian confidence rather than explicitly maximizing global coverage. In multi-room layouts this can bias selection toward local refinement and limit traversal into new rooms under a fixed flight-time budget, resulting in lower coverage in our evaluation.

\begin{table}[t]
\centering
\footnotesize
\caption{RTH budget utilization for \textbf{Ours (w/ RTH)}. $T_{\mathrm{exp}}$ is the time spent exploring before the
RTH trigger; $T_{\mathrm{rth}}$ is the time spent returning home after the trigger.}
\label{tab:rth_budget}
\vspace{-.4em}
\setlength{\tabcolsep}{5pt}
\begin{tabular}{lccc}
\toprule
Metric & Overall & Gibson & HM3D \\
\midrule
Mean exploration time $T_{\mathrm{exp}}$ (s) & 124.6 & 110.5 & 188.2 \\
Mean RTH time $T_{\mathrm{rth}}$ (s)         & 20.3  & 17.8  & 31.3 \\
Exploration fraction $T_{\mathrm{exp}}/B$ (\%) & 79.3 & 79.4 & 78.4 \\
RTH fraction $T_{\mathrm{rth}}/B$ (\%)         & 12.2 & 12.1 & 13.0 \\
Other overhead (\%)$^\dagger$                 & 8.5  & 8.5  & 8.6 \\
\bottomrule
\vspace{-0.2cm}
\end{tabular}
\footnotesize{$^\dagger$Includes initialization scan and brief safety recovery behaviors.}
\vspace{-0.4cm}
\end{table}
\subsubsection{Return-to-Home Budget Utilization}
\label{sec:rth_budget}

Since RTH feasibility is a primary design goal, we additionally quantify how the flight-time budget is allocated between exploration and the final return segment for \textbf{Ours (w/ RTH)}. 
We define \emph{RTH success} as the MAV reaching within $r_{\mathrm{home}}=1.0$\,m of the takeoff position before the flight-time budget expires.
The other baseline methods dont have RTH criteria and are not applicable here. 
We measure $T_{\mathrm{exp}}$ as the time spent executing exploration waypoints before switching into RTH mode, and $T_{\mathrm{rth}}$ as the time spent returning to the home position after the RTH trigger. We report both absolute
times and normalized fractions ($T_{\mathrm{exp}}/B$ and $T_{\mathrm{rth}}/B$).

Table~\ref{tab:rth_budget} shows that the return segment consumes only $\approx 12$--$13\%$ of the budget on average, leaving $\approx 78$--$79\%$ for exploration; the remaining fraction corresponds to fixed and safety-related overhead (e.g., brief obstacle recovery maneuvers). This helps explain why enforcing an RTH safety margin does not degrade reconstruction performance in our experiments. 
RTH failures typically arise from execution-level blockage in cluttered regions, where the depth-triggered safety policy repeatedly performs recovery maneuvers near narrow passages and consumes the remaining budget. In a smaller number of cases, the current roadmap becomes temporarily disconnected due to incomplete online carving around doorways, so the planner cannot find a feasible return path even though the true environment remains connected.


\vspace{-0.1cm}
\subsection{Ablation Study}
We conduct an ablation study to isolate the contribution of the key components in our exploration pipeline. Each ablation removes or replaces exactly one module while keeping all other settings fixed. (i) \textbf{Greedy viewpoint selection: } Our method selects viewpoints using a lightweight two-step receding-horizon lookahead (NBP) over the top-$K$
candidates. Here, we disable NBP and select the next action greedily by the single-step score $U(n,h)$. (ii) \textbf{Free-space filtering: }
We disable the free-space proximity explained in section \ref{subsec:voronoitessellation} filter and rely only on TSDF-derived geometry (mesh and distance-to-surface) for node and edge validation. 
(iii) \textbf{Dense occupancy line-of-sight filter.} To quantify how much the LoS filter contributes to suppressing ``phantom'' behind-wall nodes, we also ablate this filter. 
\begin{table}
\centering
\footnotesize
\caption{Ablations. Cov. (\%) $\uparrow$, Err. (cm) $\downarrow$, RTH (\%) $\uparrow$.}
\label{tab:ablation_clean}
\vspace{-.4em}
\begin{tabular}{llccc}
\toprule
Dataset & Method & Cov. & Err. & RTH \\
\midrule
\multirow{4}{*}{Gibson}
& Ours (RTH)                 & 88.81 & 6.27  & 82.2 \\
& w/o NBP (Greedy)      & 86.45 & 7.96  & 81.8 \\
& w/o Free-space filter      & 85.07 & 7.76  & 66.7 \\
& w/o Dense LoS filter       & 88.00 & 6.08  & 75.6 \\
\midrule
\multirow{4}{*}{HM3D}
& Ours (RTH)                 & 66.27 & 18.14 & 90.0 \\
& w/o NBP (Greedy)      & 67.24 & 18.07 & 70.0 \\
& w/o Free-space filter      & 67.05 & 18.60 & 30.0 \\
& w/o Dense LoS filter       & 66.36 & 18.67 & 80.0 \\
\bottomrule
\end{tabular}
\vspace{-0.6cm}
\end{table}
Table~\ref{tab:ablation_clean} shows that each component contributes either to exploration efficiency (Cov/Err) or to execution robustness (RTH). Replacing NBP with greedy selection consistently reduces reconstruction quality on Gibson and lowers RTH success in more complex HM3D layouts, indicating that short-horizon lookahead helps avoid myopic dead-ends and stabilizes long-range decisions. 
Disabling the free-space proximity filter causes the largest RTH drop, especially on HM3D, because incomplete online geometry near doorways and corridors can create TSDF-derived medial-axis nodes in uncarved voids or behind walls; these nodes can pass distance checks but are not safely reachable.
Finally, removing the dense occupancy line-of-sight filter produces a smaller change in mean coverage but still reduces RTH success, consistent with its role in suppressing behind-wall candidates that otherwise trigger repeated safety stops and wasted time.
Overall, NBP improves non-myopic exploration behavior, while the free-space and LoS filters are critical for maintaining a reachable, collision-consistent roadmap in multi-room indoor scenes.

\vspace{-0.2cm}
\subsection{Computational Cost Analysis}
\vspace{-0.1cm}
We profile wall-clock overhead for mapping and planning, excluding these from the simulated flight budget. During flight, our dense 3DGS backend continuously processes RGB-D frames at $\sim$4\,FPS. Between these continuous waypoint traversals, our intermittent replanning cycles average $0.93$\,s for planning and $0.84$\,s for mapping. While ActiveGS has a lower per-step planning latency ($0.42$\,s), its myopic actions force excessive replanning, accumulating massive per-episode totals ($81.5$\,s planning, $90.1$\,s mapping). In contrast, our hybrid architecture decouples 3DGS from a lightweight topological scaffold. This enables long-range waypoint actions that collect data continuously, drastically reducing the required number of replanning interventions and yielding highly efficient episodic totals of just $16.4$\,s for planning and $12.8$\,s for mapping.

\vspace{-0.3cm}
\section{Conclusion}
\vspace{-0.1cm}
We present a budget-aware active reconstruction framework for micro aerial vehicles that targets high-fidelity 3DGS while explicitly enforcing return-to-home (RTH) feasibility. Our approach decouples rendering from navigation by optimizing a dense 3DGS map as the reconstruction output and maintaining a lightweight TSDF/occupancy scaffold for conservative collision checking and online construction of a sparse 3D Voronoi skeleton roadmap. Viewpoints are selected as (node, yaw) actions using a budget-adaptive objective and a lightweight two-step receding-horizon lookahead to reduce myopic decisions under limited flight time. 
Across photorealistic indoor simulation benchmarks, our method achieves strong reconstruction performance while maintaining safe recovery behavior, attaining high RTH success and outperforming a 2D/topological Voronoi-based planner and a surface-centric 3DGS baseline in mean coverage and/or reconstruction error under identical flight-time budgets.
Limitations include a simple depth-triggered reactive safety policy that can waste time in cluttered scenes and a roadmap update procedure that rebuilds the 3D Voronoi graph periodically, contributing non-trivial computation. Future work includes more efficient incremental/event-driven roadmap maintenance and improved execution robustness. 

\vspace{-0.1cm}
\bibliographystyle{IEEEtran}
\footnotesize
\bibliography{ref}

\end{document}